\pdfoutput=1

\documentclass[11pt]{article}

\usepackage[]{EMNLP2022}

\usepackage{times}
\usepackage{latexsym}

\usepackage{booktabs,subcaption,amsfonts,dcolumn}
\usepackage{adjustbox}

\usepackage[T1]{fontenc}

\usepackage[utf8]{inputenc}

\usepackage{microtype}

\usepackage{inconsolata}

\usepackage{array}
\usepackage{pifont}
\usepackage{tabularx}
\usepackage{adjustbox}
\usepackage{multirow}
\usepackage{enumitem}
\usepackage{xspace}
\usepackage{tcolorbox}
\usepackage{booktabs,subcaption,amsfonts,dcolumn}
\usepackage{url}
\usepackage{amsmath,amsthm,amsfonts,amssymb,bm,stmaryrd}
\usepackage{color,xcolor,colortbl}
\usepackage{CJKutf8}
\usepackage{makecell}
\usepackage{booktabs}
\usepackage{graphicx}
\usepackage{listings}

\definecolor{mygreen}{rgb}{0.64, 0.76, 0.68}
\definecolor{myyellow}{rgb}{0.98, 0.94, 0.75}
\definecolor{mygreen}{rgb}{0.68, 0.85, 0.9}
\definecolor{myblue}{rgb}{0.82, 0.94, 0.75}
\definecolor{mypurple}{RGB}{224, 65, 245}
\definecolor{myorange}{RGB}{209, 136, 17}

\title{Towards Realistic Low-resource Relation Extraction:\\ A Benchmark with Empirical Baseline Study}

\author{
Xin Xu\textsuperscript{\rm 1,2}\footnotemark[1], 
Xiang Chen\textsuperscript{\rm 1,2}\thanks{$\quad$ Equal contribution and shared co-first authorship.}, 
Ningyu Zhang\textsuperscript{\rm 1,2\thanks{\quad Corresponding author.}}, 
Xin Xie\textsuperscript{\rm 1,2}, 
Xi Chen\textsuperscript{\rm 3}, 
Huajun Chen\textsuperscript{\rm 1,2}  \\
\textsuperscript{\rm 1}Zhejiang University \& AZFT Joint Lab for Knowledge Engine\\
\textsuperscript{\rm 2}Hangzhou Innovation Center, Zhejiang University,
\textsuperscript{\rm 3}Tencent\\
 \{xxucs@zju.edu.cn, xiang\_chen, xx2020, huajunsir, zhangningyu\}@zju.edu.cn, \\
 jasonxchen@tencent.com\\
 \url{https://zjunlp.github.io/project/LREBench}
}

\begin{document}
\maketitle
\begin{abstract}
This paper presents an empirical study to build relation extraction systems in low-resource settings. Based upon recent pre-trained language models, we comprehensively investigate three schemes to evaluate the performance in low-resource settings: $(i)$ different types of prompt-based methods with few-shot labeled data;  $(ii)$ diverse balancing methods to address the long-tailed distribution issue; $(iii)$ data augmentation technologies and self-training to generate more labeled in-domain data. We create a benchmark with 8 relation extraction (RE) datasets covering different languages, domains and contexts and perform extensive comparisons over the proposed schemes with combinations. Our experiments illustrate: $(i)$ Though prompt-based tuning is beneficial in low-resource RE, there is still much potential for improvement, especially in extracting relations from cross-sentence contexts with multiple relational triples; $(ii)$ Balancing methods are not always helpful for RE with long-tailed distribution;  $(iii)$ Data augmentation complements existing baselines and can bring much performance gain, while self-training may not consistently achieve advancement to low-resource RE\footnote{Code and datasets 
are in \url{https://github.com/zjunlp/LREBench}.}.

\end{abstract}

\section{Introduction}

Relation Extraction (RE) aims to extract relational facts from the text and plays an essential role in information extraction \cite{zhang2022deepke}.
The success of neural networks for RE has been witnessed in recent years; however, open issues remain as they still depend on the number of labeled data in practice.
For example, \citet{DBLP:conf/emnlp/HanYLSL18} found that the model performance drops dramatically as the number of instances for one relation decreases, e.g., for long-tail.
An extreme scenario is few-shot RE, where only a few support examples are given.
This motivates a \textbf{Low-resource RE} (\textbf{LRE}) task where annotations are scarce \cite{DBLP:conf/emnlp/BrodyWB21}.

\begin{figure}
    \centering
    \includegraphics[width=0.49\textwidth]{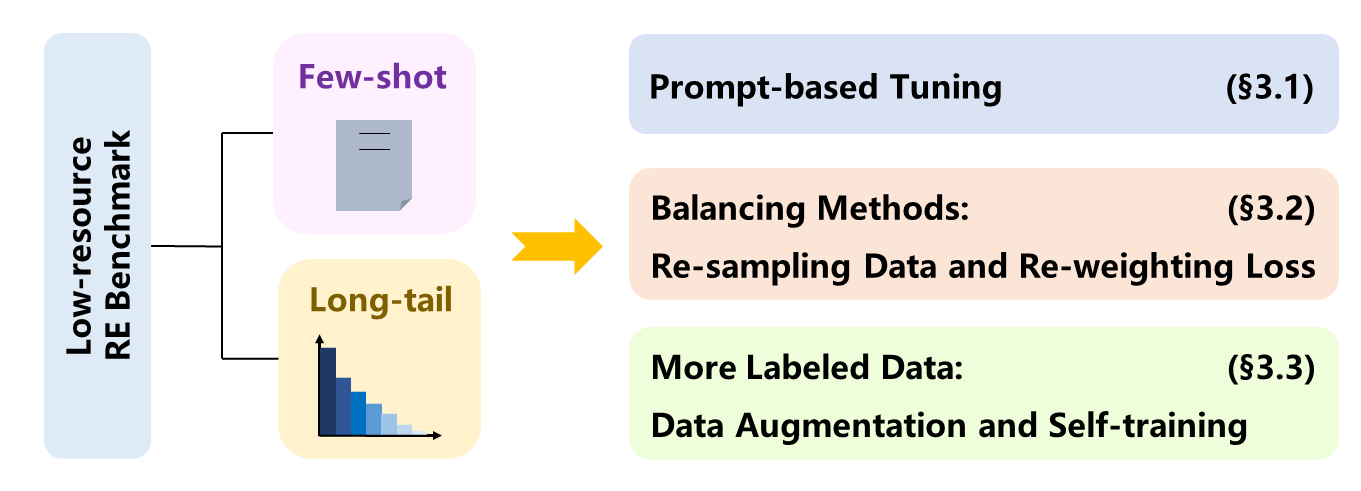}
    \caption{An overview of methods studied in our paper.}
    \label{motivation}
\end{figure}

Many efforts are devoted to improving the generalization ability beyond learning directly from limited labeled data.
Early, \citet{DBLP:conf/acl/MintzBSJ09} proposes distant supervision for RE, which leverages facts in KG as weak supervision to obtain annotated instances. 
\citet{DBLP:conf/wacv/RosenbergHS05,DBLP:conf/naacl/LiuFTCZHG21,DBLP:conf/emnlp/HuZYLLWY21} try to assign pseudo labels to unlabeled data and leverage both pseudo-labeled data and gold-labeled data to improve the generalization capability of models iteratively.
Some studies apply meta-learning strategies to endow a new model with the ability to optimize rapidly or leverage transfer learning to alleviate the data-hungry issue \cite{DBLP:conf/aaai/GaoH0S19,DBLP:conf/coling/YuZDYZC20,DBLP:conf/coling/LiWZZYC20,DBLP:conf/acl/DengZLHTCHC20}. 
Other studies \cite{DBLP:conf/naacl/ZhangDSWCZC19} focus on the long-tailed class distribution, especially in tail classes that only allow learning with a few instances.
With the prosperity of the pre-trained language models (PLMs),  the pre-train -- fine-tune paradigm has become standard for natural language processing (NLP), leading to a tremendous increase in LRE performance.
More recently, a new methodology named prompt learning has made waves in the community by demonstrating astounding few-shot capabilities on LRE \cite{han2021ptr,DBLP:conf/www/ChenZXDYTHSC22}.

In this work, we benchmark more realistic scenarios on diverse datasets for low-resource RE, in which models have to handle \textbf{both extreme few-shot instances and long-tailed distribution, and can also make use of data augmentation or unlabeled in-domain data without cross-validation} \cite{DBLP:conf/nips/PerezKC21}.
These settings are appealing as: 
$(i)$ Such models mirror deployment in applied settings;
$(ii)$ Few-shot settings are realistic with long-tailed distribution;
$(iii)$ Diverse datasets cover different languages (Chinese and English), domains (general, scientific), and contexts (one or more sentences with single or multiple relational triples). 
 
Specifically, we focus on improving the generalization ability from three directions shown in Figure \ref{motivation}.
Instead of using limited few-shot data, 
we create different types of prompts for RE and empirically analyze low-resource performance.
We further implement many popular balancing methods for long-tailed distribution, which can mitigate performance decay in instance-scarce (tail) classes.
We also leverage more generated training instances by data augmentation and self-training in conjunction with the limited labeled data. 

Our contributions include: 
$(i)$  We present the \textbf{first} systematic study for low-resource RE, an important problem in information extraction, by investigating three distinctive schemes with combinations.
$(ii)$ We conduct extensive comparisons with in-depth analysis on 8 RE datasets and report empirical results with insightful findings.
$(iii)$ We release both the data and the source code of these baselines as an open-sourced testbed for future research purposes.

To shed light on future research on low-resource RE, our empirical analysis suggests that: 
$(i)$ Previous state-of-the-art methods in the low-resource setting still struggle to obtain better performance than that in the fully-supervised setting (Cross-sentence LRE is extremely challenging), which indicates that there is still much room for low-resource RE.
$(ii)$ Balancing methods may not always benefit low-resource RE. 
The long-tailed issue can not be ignored, and more studies should be focused on model development.
$(iii)$ With some simple data augmentation methods, better performance can be achieved, highlighting opportunities for future improvements on low-resource RE.

\begin{figure*}[t!]
	\vspace{-0mm}\centering
	\begin{tabular}{c c}
		\hspace{-4mm}
		\includegraphics[height=4cm,width=7.7cm]{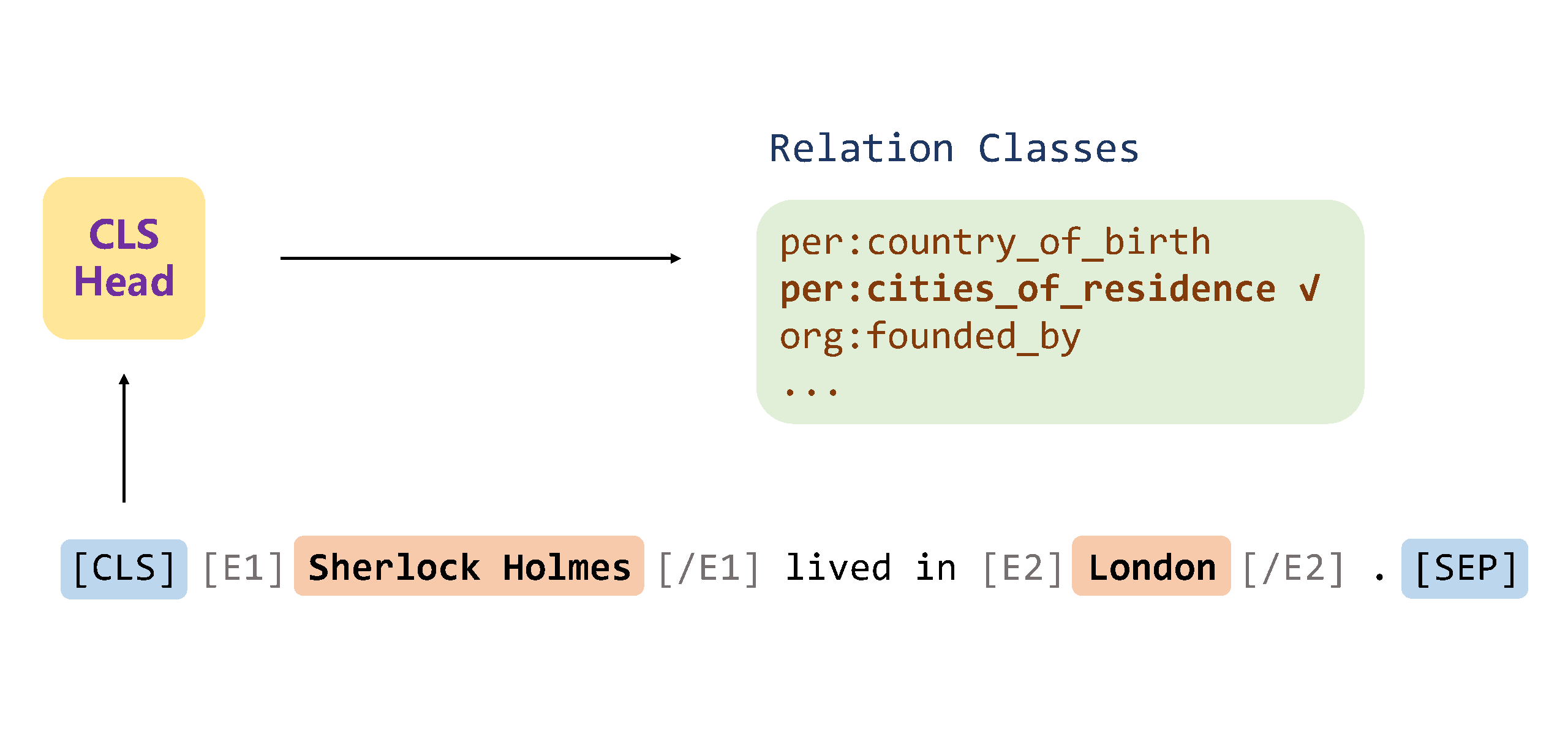} & 
		\hspace{-3mm}
		\includegraphics[height=4cm,width=8.2cm]{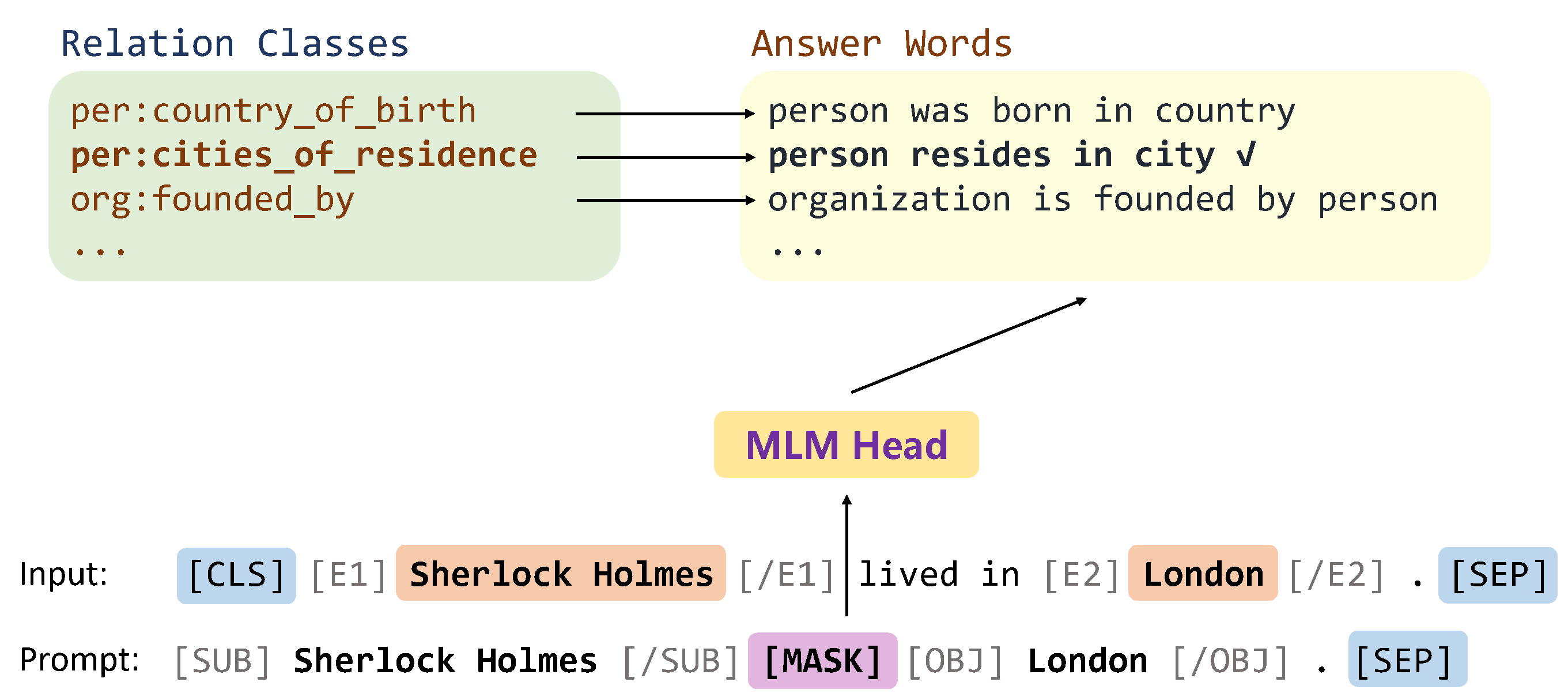} \\
		\vspace{0mm}
		\hspace{-6mm}
		(a) Standard Fine-tuning \hspace{3mm} & 
		(b) Prompt-based Tuning   \vspace{2mm}\\ 
		\includegraphics[height=6.8cm,width=6.9cm]{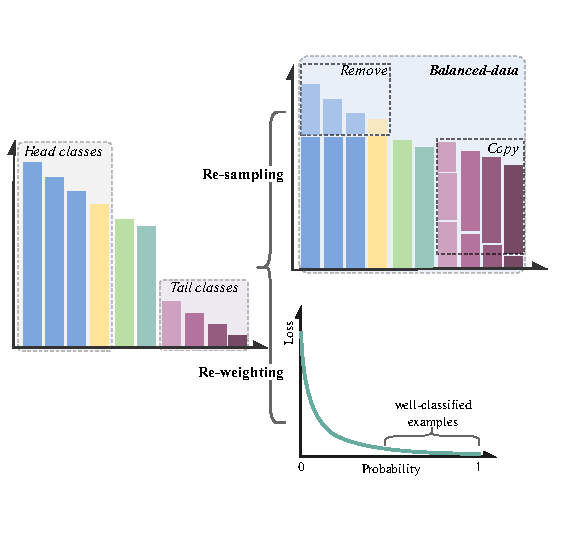} & 
		\hspace{3mm}
		\includegraphics[height=6.6cm,width=7.4cm]{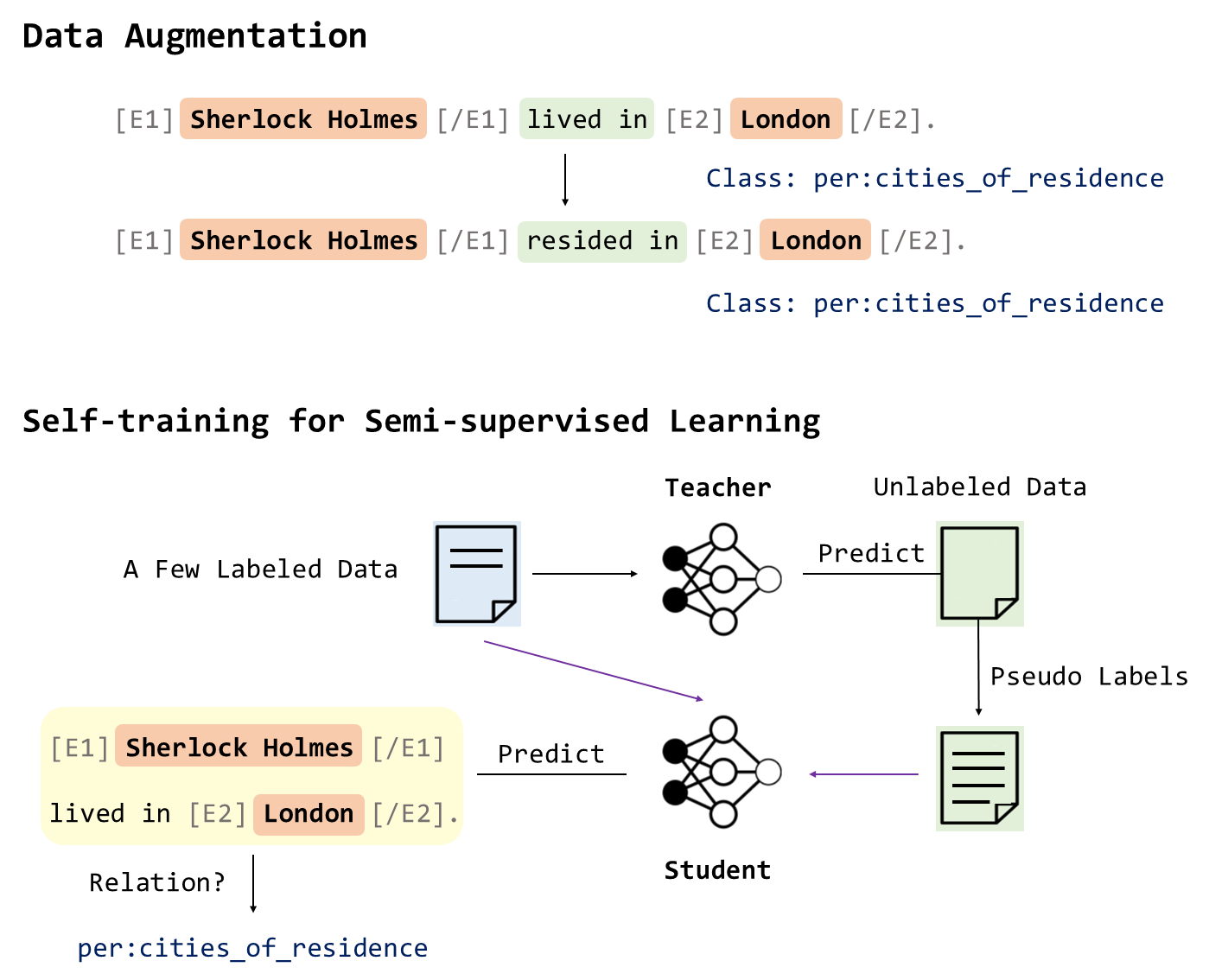} \\
		(c) Balancing Methods  \vspace{0mm} & 
		(d) Leveraging More Instances \hspace{-0mm} \\ 
	\end{tabular}
	\vspace{-0mm}
	\caption{
Illustrations of different methods used in our low-resource RE benchmark.
(a) A standard RE pipeline of fine-tuning a PLM such as BERT and RoBERTa (\S \ref{ft}).
(b) Prompt-based tuning,  which concatenates the original input with the prompt template to predict $\texttt{[MASK]}$ by an MLM head and then injects the predicted answer words to the corresponding class sets (\S \ref{prompt}).
(c) Two balancing methods, re-sampling data and re-weighting losses, to address the long-tailed issue (\S \ref{balance}). 
(d) Levering more instances with data augmentation and self-training (\S \ref{sec:DA and ST}). 
	 }
	\vspace{-2mm}
	\label{fig:RE}
\end{figure*}

\section{Background on Low-resource RE}
\subsection{Low-resource RE}
RE is a classification task that aims to assign relation labels to entity pairs in given contexts. 
Formally, in a RE dataset denoted as $\mathcal{D} = \{\mathbf{X}, \mathbf{Y}\}$, $\mathbf{X}$ is the set of texts and $\mathbf{Y}$ is the set of relation labels. Given a text $x = \{w_1, w_2, \ldots, w_s, \ldots, w_o,\ldots, w_{\mid x \mid} \}$, where $x \in \mathbf{X}$, RE aims to predict the semantic relation $y_x \in \mathbf{Y}$ holding between the subject entity $w_s$ and the object entity $w_o$.
Conventional RE systems are trained in the standard supervised learning regime, where large amounts of labeled examples are required.
Nevertheless, owing to various languages, domains, and the cost of human annotation, there is commonly a very small number of labeled examples in real-world applications. 
Thus, traditional supervised learning with few-shot labeled data struggle to achieve satisfactory performance \cite{schick-schutze-2021-exploiting}.
Consequently, a challenging task, \emph{low-resource RE}, has emerged.

\subsection{Fine-tuning PLMs for RE}
\label{ft}
A typical baseline method for RE is to fine-tune a PLM $\mathcal{M}$ as shown in Figure \ref{fig:RE}(a).
Firstly, the tokenizer of $\mathcal{M}$ converts the text $x$ into the input tokens of $\mathcal{M}$, such as $ \texttt{[CLS]} x_{\text{token}} \texttt{[SEP]}$, and then encodes tokens into the corresponding hidden vectors, such as $\mathbf{h} = \{ \mathbf{h}_{\texttt{[CLS]}}, \mathbf{h}_1, \mathbf{h}_2, \ldots, \mathbf{h}_s, \ldots, \mathbf{h}_o, \ldots, \mathbf{h}_{\texttt{[SEP]}} \}$. 
Then, a \texttt{[CLS]} head is used to compute the probability distribution over the class set $\mathbf{Y}$ with the softmax  $p(\cdot | x) = \texttt{Softmax}(\mathbf{Wh}_{\texttt{[CLS]}}+\mathbf{b})$, where $\mathbf{W}$ is a set of learnable weight parameters randomly initialized at the start of fine-tuning, $\mathbf{h}_{\texttt{[CLS]}}$ is the hidden vector of \texttt{[CLS]} and $\mathbf{b}$ is the learnable bias. 
All learnable parameters are fine-tuned by minimizing the cross-entropy loss over $p(y_x|x)$ on $\mathcal{D}$.
Nevertheless, conventional supervised fine-tuning may over-fit a few training examples and perform poor generalization ability over test sets when encountering the low-resource RE task.

\section{Methods for Low-resource RE}
In this paper, we conduct a comprehensive empirical study with three distinctive schemes against difficulty in low-resource RE: PLMs-based prompt-based tuning, balancing long-tailed data and leveraging more instances, as shown in Figure \ref{fig:RE}.

\subsection{Prompting for Few-shot Instances}
\label{prompt}
To address the low-resource issue of data sparsity for RE, we firstly analyze \textbf{prompting methods}.
Unlike standard fine-tuning, prompt-based tuning reformulates classification tasks as cloze-style language modeling problems and predicts answer words, denoted as $\mathbf{V}$, through the masked language model (MLM) head.
Specifically, $\mathcal{T}_{\text{prompt}}$ converts every instance $x$ into a prompt input $x_{\text{prompt}} = \mathcal{T}_{\text{prompt}}(x)$, in which there is at least one \texttt{[MASK]} for $\mathcal{M}$ to fill with right answer words $v \in \mathbf{V}$.
Meanwhile, a verbalizer connects relation labels with answer words via an injective mapping $\mathcal{\gamma}: \mathbf{Y} \rightarrow \mathbf{V}$.
With the aforementioned functions, we can formalize the probability distribution over $\mathbf{Y}$ with the probability distribution over $\mathbf{V}$ at the masked position \cite{DBLP:journals/corr/abs-2109-13532}:
\begin{align}
\small
    \begin{split}
        P(y_x|x) &= P(\texttt{[MASK]}=\gamma(y_x)|x_{prompt}) \\
        &= Softmax(\mathbf{W}_{lm} \cdot \mathbf{h}_{\texttt{[MASK]}})
    \end{split}
\end{align}
where $\mathbf{W}_{lm}$ is a set of parameters of the PLM head. 
 
Note that the main difference between various prompt-based tuning methods lies in the design of the prompt template and verbalizer.
Thus, we benchmark different kinds of prompting methods in low-resource RE to empirically investigate their performance.
For the prompt template, given the input $x$, the first choice is manually designing the template. 
We utilize the natural language or task schema to formulate different prompt templates.
Formally, we have: \\ \\
\scalebox{1}{
\small
\begin{tabular}{l}
     {\bf Template Prompt:}\\\texttt{[CLS]} $x$.\ \texttt{[SEP]}\ The relation between ${\tt [sub]}$\ and ${\tt [obj]}$\\ is \texttt{[MASK]}. \texttt{[SEP]}  \\ \\
     {\bf Schema Prompt:}\\\texttt{[CLS]} $x$.\ \texttt{[SEP]}\ [ ${\tt [sub]}$ | ${\tt [obj]}$ ] relation: \texttt{[MASK]}. \texttt{[SEP]} \\
\end{tabular}}
\\ \\
where \texttt{<sub>} is the head entity mention and \texttt{<obj>} is the tail entity mention.
Since there exists rich semantic knowledge within relation labels and structural knowledge implications among relational triples, we also benchmark previous studies such as \textit{PTR} \cite{han2021ptr} and \textit{KnowPrompt} \cite{DBLP:conf/www/ChenZXDYTHSC22} which incorporates relational knowledge into prompt-based tuning as shown in Figure \ref{fig:RE}(b).

\subsection{Balancing for Long-tailed Distribution}
\label{balance}
Learning with long-tailed data, where the number of instances in each class highly varies, is a common challenge in low-resource RE because instance-rich (head) classes predominate the training procedure.
Note that the learnable parameters of the trained model prefer to perform better in these head classes and worse in less frequent (tail) classes \cite{DBLP:conf/iclr/KangXRYGFK20}. 
To address this issue, we explore two balancing methods: re-sampling data and re-weighting losses for low-resource RE.

\paragraph{Re-sampling Data}
We re-sample RE datasets to balance the data distribution. 
For example, the tail classes can be over-sampled by adding copies of data, and the head classes can be under-sampled by removing data, as shown in Figure \ref{fig:RE}(c).
Specifically, we use a toolkit\footnote{ \url{https://github.com/ufoym/imbalanced-dataset-sampler}}, which can estimate the sampling weights automatically when sampling from imbalanced data to obtain datasets with the nearly balanced distribution.

\paragraph{Re-weighting Loss} 
We utilize various re-weighting losses, assigning different weights to different training instances for each class.
For instance, \textit{DSC Loss} \cite{DBLP:conf/acl/LiSMLWL20} attaches similar importance to false positives and false negatives.
\textit{Focal Loss} \cite{DBLP:journals/pami/LinGGHD20} balances the sample-wise classification loss for model training by down-weighing easy samples. 
\textit{GHM Loss} \cite{10.1609/aaai.v33i01.33018577} applies a gradient harmonizing mechanism, making the model ignore outliers to conquer the disharmony in classification. 
\textit{LDAM Loss} \cite{DBLP:conf/nips/CaoWGAM19} expands the decision boundaries of few-shot classes.

\subsection{Leveraging More Instances via Data Augmentation and Self-training}
\label{sec:DA and ST}
It is also beneficial to leverage more instances to address the low-resource issue.
We conduct data augmentation and also leverage unlabeled in-domain data via self-training, as shown in Figure \ref{fig:RE}(d).

Data augmentation (DA) automatically generates more labeled instances based on only a few labeled instances.
For example, we utilize token-level augmentation, which changes or inserts words and phrases in a sentence to generate augmented text remaining with the same labels as the original text.
In this work, we apply three DA methods for English RE datasets to substitute words in training sets based on \textbf{WordNet's synonyms}, \textbf{TF-IDF similarity} and the \textbf{contextual word embedding} implemented by \textit{nlpaug}\footnote{\url{https://github.com/makcedward/nlpaug}}.
And we replace words with their synonyms via \textit{nlpcda}\footnote{\url{https://github.com/425776024/nlpcda}} to augment Chinese RE samples.
We further analyze different types of augmentation objects in RE regarding contexts, entities, and both of them.

\begin{table*}[h]
  \centering
  \scalebox{0.7}{
    \begin{tabular}{c|c|c|c|c|c|c|c|c}
    \toprule
    Datasets & \textbf{SemEval} & \textbf{TACREV} & \textbf{Wiki80}*  & \textbf{SciERC} & \textbf{ChemProt} & \textbf{DialogRE} & \textbf{DuIE2.0} (cn) & \textbf{CMeIE} (cn) \\
    \midrule
    Domain & General & General & Encyclopedic  & Scientific & Biochemical & Dialogue & General & Medical \\
    \# Train & 6.5k & 68.1k & 12.0k  & 3.2k & 19.5k & 6.0k & 153k & 34k \\
    \# Test & 2.7k & 15.5k & 5.6k  & 974 & 16.9k & 1.9k & 18k & 8.7k \\
    \# Relation Class & 19 & 42 & 80 & 7 & 14 & 37 & 48 & 44 \\
    MS / MT & $\times$ / $\times$ & $\times$ / $\times$ & $\times$ / $\times$ & $\checkmark$ / $\checkmark$ & $\checkmark$ / $\checkmark$ & $\checkmark$ / $\checkmark$ & $\times$ / $\checkmark$ & $\checkmark$ / $\checkmark$ \\
    \bottomrule
    \end{tabular}
    }
 \caption{
 Statistics on the 8 public RE datasets selected for evaluation in \textbf{LREBench}.
 \textbf{MS} indicates if datasets contain instances with multiple sentences in one text, and \textbf{MT} indicates if one text in these datasets can be related to multiple relational triples.
 "*'' means that we \textbf{re-sample} and convert Wiki80 into long-tailed distribution through an exponential function since its original distribution is exactly balanced.
 "cn'' represents datasets with Chinese.}
  \label{tab:dataset}
\end{table*}

Since substantial easily-collected unlabeled data are also leveraged in this work for low-resource RE, we conduct self-training, a classical, intuitive and straightforward semi-supervised learning method.
Specifically, we train a model with labeled data and then expand the labeled set according to the most confident predictions (a.k.a. pseudo labels) on unlabeled data.
We combine the data with gold and pseudo labels to obtain the final RE model.
The details of the whole self-training pipeline are described in Appendix \ref{app:self-train}.

\section{Benchmark Design}

In this paper, we provide a comprehensive empirical study for low-resource RE and design the \textbf{LREBench} (\textbf{L}ow-resource \textbf{R}elation \textbf{E}xtraction \textbf{Bench}mark) to evaluate various methods.
In the following section, we will detail the datasets chosen for experiments and the reproducibility of all baselines mentioned above.

\subsection{Datasets Selection}
As shown in Table \ref{tab:dataset}, we select 8 RE datasets to evaluate baselines in low-resource settings, covering various domains: SemEval 2010 Task 8\footnote{\url{https://github.com/zjunlp/KnowPrompt/tree/master/dataset/semeval}} \cite{hendrickx-etal-2009-semeval}, TACREV\footnote{\url{https://github.com/DFKI-NLP/tacrev}} \cite{alt-etal-2020-tacred}, DialogRE\footnote{\url{https://dataset.org/dialogre/}} \cite{yu-etal-2020-dialogue} and DuIE2.0\footnote{\url{https://www.luge.ai/\#/luge/dataDetail?id=5}} \cite{10.1007/978-3-030-32236-6_72} on the general domain, Wiki80\footnote{\url{https://github.com/thunlp/OpenNRE/blob/master/benchmark/download_wiki80.sh}} \cite{han-etal-2019-opennre} on the encyclopedic domain, ChemProt\footnote{\url{https://github.com/ncbi-nlp/BLUE_Benchmark}} \cite{DBLP:conf/bionlp/PengYL19} on the biochemical domain, SciERC\footnote{\url{http://nlp.cs.washington.edu/sciIE/}} \cite{luan-etal-2018-multi} on the scientific domain, and CMeIE\footnote{\url{https://tianchi.aliyun.com/dataset/dataDetail?dataId=95414}} \cite{zhang-etal-2022-cblue} on the medical domain.
Except for frequently-used English datasets, we select Chinese datasets, such as DuIE2.0 and CMeIE.
Besides, the SciERC, ChemProt, DialogRE, and CMeIE datasets contain the situation where multiple sentences are in one instance, which is for cross-sentence RE and more challenging than single-sentence RE in SemEval, TACREV and Wiki80.

For simplicity, we provide a unified input-output format for all datasets in the low-resource setting\footnote{We utilize a unified json format for evaluation, and it is straightforward to adapt to other datasets.}.
Specifically, each instance in LREBench consists of one text and one relational triple (one head entity and one tail entity in the text and the corresponding relation between them).
For those datasets with instances having one text related to multiple relational triples, such as ChemProt, SciERC, DialogRE, DuIE2.0 and CMeIE, we follow \citet{DBLP:conf/naacl/ZhongC21} to place such a text to multiple instances with only one relational triple.
In this way, we can utilize a unified input-output format for widespread models.

We conduct experiments in three settings with different proportions of training data to simulate different resource levels: 8-shot, 10\% and 100\%.
For the 8-shot setting, we sample 8 instances for each relation category in the training and test sets\footnote{If there are less than 8 instances in one relation class, we delete all instances of this class.}.
For the 10\% and 100\% settings, we sample 10 percent of the training set and use the whole training set, respectively.
Since fine-tuning on small datasets can suffer from instability and results may change dramatically given a new split of data \cite{gao-etal-2021-making}, \textbf{we sample all training datasets 5 times randomly in 8-shot and 10\% settings and measure their average performance in experiments}.
Also, we follow the same sampling strategy in the re-sampling long-tailed data method and data augmentation methods to obtain a fair comparison.

\subsection{Reproducibility}
\paragraph{Methods} Throughout our experiments, we employ $\mathcal{M} = $ \textit{RoBERTa-large} \cite{DBLP:journals/corr/abs-1907-11692} for SemEval, TACREV, Wiki80 and DialogRE, Chinese \textit{RoBERTa-large} \cite{cui-etal-2020-revisiting} for DuIE2.0 and CMeIE, and \textit{BioBERT-large} \cite{lee2020biobert} for ChemProt and SciERC from \textit{HuggingFace}\footnote{\url{https://huggingface.co/}} as the backbone network (detailed in Appendix \ref{app:setting}).
For each method, we investigate the following three schemes in different settings for the comparative empirical study, as shown in Table \ref{tab:results}: 
$(i)$ \textbf{Normal} is the general scheme with the PLM for low-resource relation extraction, in which we evaluate with 8-shot, 10\% and 100\% settings. 
$(ii)$ \textbf{Balance} refers to balancing methods in \S \ref{balance} for long-tailed data distribution with 10\% and 100\% settings. 
We list the best performance among all balancing methods for each dataset in Table \ref{tab:results} and detailed results in Table \ref{tab:balance}.
$(iii)$ \textbf{Data augmentation (DA)} methods are applied to 10\% training sets. 
We list the best performance among all DA methods in Table \ref{tab:results} and all performance in Table \ref{tab:DA}.
We also conduct \textbf{self-training (ST)} that firstly trains a teacher $\mathcal{M}$ on 10\% training data and then tags the rest 90\% training data with pseudo labels by $\mathcal{M}$. 
Both gold-labeled and pseudo-labeled data are used to obtain a final student RE model as introduced in \S \ref{sec:DA and ST}.

\begin{table*}[t!]
  \centering
  \scalebox{0.7}{
    \begin{tabular}{lc|ccc|cc|cccccc|cc|cc}
    \toprule
    \multicolumn{1}{l}{\multirow{3}[4]{*}{\textbf{Dataset}}} & \multicolumn{1}{c}{\multirow{3}[4]{*}{\textbf{Metric}}} & \multicolumn{7}{c}{\textbf{Fine-Tune}} &   & \multicolumn{7}{c}{\textbf{Prompt}} \\
\cmidrule{3-9}\cmidrule{11-17}    \multicolumn{1}{l}{} & \multicolumn{1}{c}{} & \multicolumn{3}{c}{\textbf{Normal}} & \multicolumn{2}{c}{\textbf{Balance}} & \textbf{DA} & \textbf{ST}&  & \multicolumn{3}{c}{\textbf{Normal}} & \multicolumn{2}{c}{\textbf{Balance}} & \textbf{DA} & \textbf{ST} \\
    \multicolumn{1}{l}{} & \multicolumn{1}{c}{} & 8-shot & 10\% & \multicolumn{1}{c}{100\%} & 10\% & \multicolumn{1}{c}{100\%} & 10\% & 10\% &  & 8-shot & 10\% & \multicolumn{1}{c}{100\%} & 10\% & \multicolumn{1}{c}{100\%} & 10\% & 10\% \\
\cmidrule{1-9}\cmidrule{11-17}    \multirow{2}[2]{*}{SemEval} & MaF1 & 2.69 & 34.63 & 81.88 & 41.84 & 82.44 & 69.84 & 60.10 &  & 48.54 & 44.71 & 83.40 & 54.54 & \color{blue}{83.20} & 71.73 & 63.55 \\
      & MiF1 & 9.70 & 54.61 & 89.10 & 58.26 & 89.44 & 78.98 & 74.12 &  & 54.55 & 69.90 & 90.01 & 76.53 & 92.31 & 83.54 & 76.81 \\
\cmidrule{1-9}\cmidrule{11-17}    \multirow{2}[2]{*}{TACREV} & MaF1 & 1.02 & 47.32 & 63.41 & 48.64 & \color{blue}{63.38} & 50.68 & 48.84 &  & 29.46 & 61.40 & 67.08 & 63.09 & 69.63 & 62.20 & \color{mypurple}{7.32}\\
      & MiF1 & 1.76 & 65.43 & 71.68 & 67.19 & 73.86 & 65.99 & 66.89 &  & 30.88 & 77.00 & 78.30 & \color{blue}{76.25} & 81.41 & \color{myorange}{76.90} & \color{mypurple}{32.93} \\
\cmidrule{1-9}\cmidrule{11-17}    \multirow{2}[2]{*}{Wiki80} & MaF1 & 37.89 & 37.82 & 71.31 & 44.37 & 73.36 & 49.40 & \color{mypurple}{37.47} &  & 75.11 & 60.67 & 82.79 & 63.99 & 83.72 & 63.40 & 60.86\\
      & MiF1 & 44.85 & 46.50 & 72.82 & 49.74 & 74.20 & 55.00 & \color{mypurple}{45.91} &   & 76.34 & 64.86 & 82.96 & 67.86 & 83.86 & 66.96 & 65.04 \\
\midrule \midrule

\multirow{2}[2]{*}{SciERC} & MaF1 & 10.41 & 10.31 & 83.41 & \color{blue}{10.11} & \color{blue}{81.17} & 30.09 & 31.48 & & 23.26 & 51.71 & \color{red}{83.27} & 60.55 & 84.83 & 65.98 & 56.94\\
      & MiF1 & 39.12 & 54.66 & 89.12 & 54.72 & \color{blue}{87.78} & 61.79 & 64.07 &   & \color{red}{22.07} & 74.00 & \color{red}{89.01} & 76.90 & 90.04 & 79.92 & 76.32\\
\cmidrule{1-9}\cmidrule{11-17}
\multirow{2}[2]{*}{ChemProt} & MaF1 & 2.18 & 27.96 & 47.35 & 33.38 & 47.35 & 36.31 & 30.67 &  & 6.17 & 36.43 & \color{red}{47.16} & 38.99 & \color{blue}{47.07} & 37.44 & \color{mypurple}{33.62}\\
      & MiF1 & 8.93 & 49.20 & 68.81 & 54.98 & \color{blue}{68.77} & 56.58 & 54.17 &  & \color{red}{8.65} & 56.96 & 69.14 & 57.28 & \color{blue}{69.12} & 58.26 & \color{mypurple}{53.55}\\
\cmidrule{1-9}\cmidrule{11-17}    \multirow{2}[2]{*}{DialogRE} & MaF1 & 1.13 & 2.17 & 25.31 & 5.84 & 27.28 & 9.74 & \color{mypurple}{0.00} &  & 44.96 & 45.51 & 64.49 & 46.22 & 71.73 & 49.47 & \color{mypurple}{34.70} \\
      & MiF1 & 3.92 & 23.37 & 41.52 & 24.53 & \color{blue}{41.24} & 27.40 & \color{mypurple}{0.00} & & 45.70 & 54.16 & 73.66 & 55.65 & \color{blue}{73.52} & 57.53 & \color{mypurple}{46.54}\\
\cmidrule{1-9}\cmidrule{11-17}    \multirow{2}[2]{*}{DuIE2.0} & MaF1 & 36.62 & 90.46 & 95.01 & 92.91 & 96.00 & 91.47 & \color{mypurple}{89.27} &  & 80.31 & 93.48 & 95.73 & 93.70 & 96.01 & 93.66 & \color{mypurple}{90.49}\\
      & MiF1 & 39.00 & 94.42 & 96.22 & 94.46 & \color{blue}{96.13} & 94.46 & \color{mypurple}{93.81} &  & 82.14 & 95.09 & 96.43 & 95.23 & 96.44 & 95.11 & \color{mypurple}{93.35}\\
\cmidrule{1-9}\cmidrule{11-17}    \multirow{2}[2]{*}{CMeIE} & MaF1 & 13.68 & 62.30 & 84.37 & 67.22 & 86.31 & 63.82 & \color{mypurple}{58.46} &  & 36.54 & 67.59 & 86.42 & 67.84 & 86.68 & 69.95 & \color{mypurple}{65.79}\\
      & MiF1 & 17.05 & 79.82 & 90.48 & 80.43 & 90.56 & 80.14 & \color{mypurple}{78.92} &  & 38.02 & 83.38 & 92.08 & 83.40 & 92.14 & 83.71 & \color{mypurple}{81.26}\\
    \bottomrule
    \end{tabular}}
    \caption{F1 Scores (\%) on 8 datasets with various sizes of training data in different methods for the low-resource scenario. 
    \textit{MaF1} and \textit{MiF1} mean Macro F1 Score (\%) and Micro F1 Score (\%) respectively. 
    \textbf{\textit{Normal}} means the standard PLM fine-tuning method and \textbf{\textit{Prompt}} means prompt-based tuning implemented by \textit{KnowPrompt}. \textbf{\textit{Balance}} represents balancing methods for long-tailed data. 
    \textbf{\textit{DA}} is data augmentation. 
    \textbf{\textit{ST}} refers to self-training with unlabeled in-domain data. 
    Results colored with \textcolor{red}{red} means prompt-based tuning works worse than fine-tuning between two \textit{Normal} columns.
    \textcolor{blue}{blue}, \textcolor{myorange}{orange}, and \textcolor{mypurple}{purple} results indicates the performance of balancing methods, data augmentation and self-training is poorer than the \textit{Normal} method in the same setting.}
    \label{tab:results}
\end{table*}

\paragraph{Training and Evaluation}
We only train models on training sets \textbf{without validation on development sets} to ensure true few-shot learning with limited labeled data.
For all training data sizes, we set the training epoch = 10 following \citet{huang-etal-2021-shot}.
Except for re-weighting losses for addressing the long-tailed problem, the cross-entropy loss is used in all training processes.
Since the performance of head and tail classes varies a lot, we use both Macro F1 and Micro F1 together as the evaluation metrics. 
Implementation details can be found in Appendix \ref{app:exper}.

\section{Results and Discussions}

\subsection{Main Results}
We leverage the basic PLM fine-tuning code from \textit{OpenNRE}\footnote{\url{https://github.com/thunlp/OpenNRE}} \cite{han-etal-2019-opennre} and the state-of-the-art prompt-based RE method \textit{KnowPrompt} \cite{DBLP:conf/www/ChenZXDYTHSC22} to conduct extensive experiments across 8 datasets in various methods and settings.
The results of the main experiments are shown in Table \ref{tab:results}, which illustrates the following findings:

\textbf{Finding 1: Prompt-based tuning largely outperforms standard fine-tuning for RE, especially more effective in the low-resource scenario.}
The comparison between the results of standard fine-tuning and prompt-based tuning indicates that prompts can provide task-specific information and bridge the pre-train -- fine-tune gap, thus, empowering PLMs in low-resource RE.

\textbf{Finding 2: Though balancing methods obtain advancement with long-tailed distribution, they may still fail on challenging RE datasets}, such as ChemProt, DialogRE and DuIE2.0.
By comparing Macro F1 Scores of the \textit{Balance} columns and \textit{Normal} columns, \textcolor{blue}{blue} (bad) results illustrate that balancing methods are affected by complexity of long contexts with multiple sentences and relational triples.

\textbf{Finding 3:}
\textbf{Data augmentation achieves much gain on RE and sometimes even better performance than prompt-based tuning}, such as on SemEval, according to the difference between two pairs of \textit{DA} columns and \textit{Normal} columns in the 10\% setting.
More data generated through DA methods are complementary with other baselines, boosting the performance.

\begin{figure*}[t!]
    \centering
    \includegraphics[width = 0.78\linewidth]{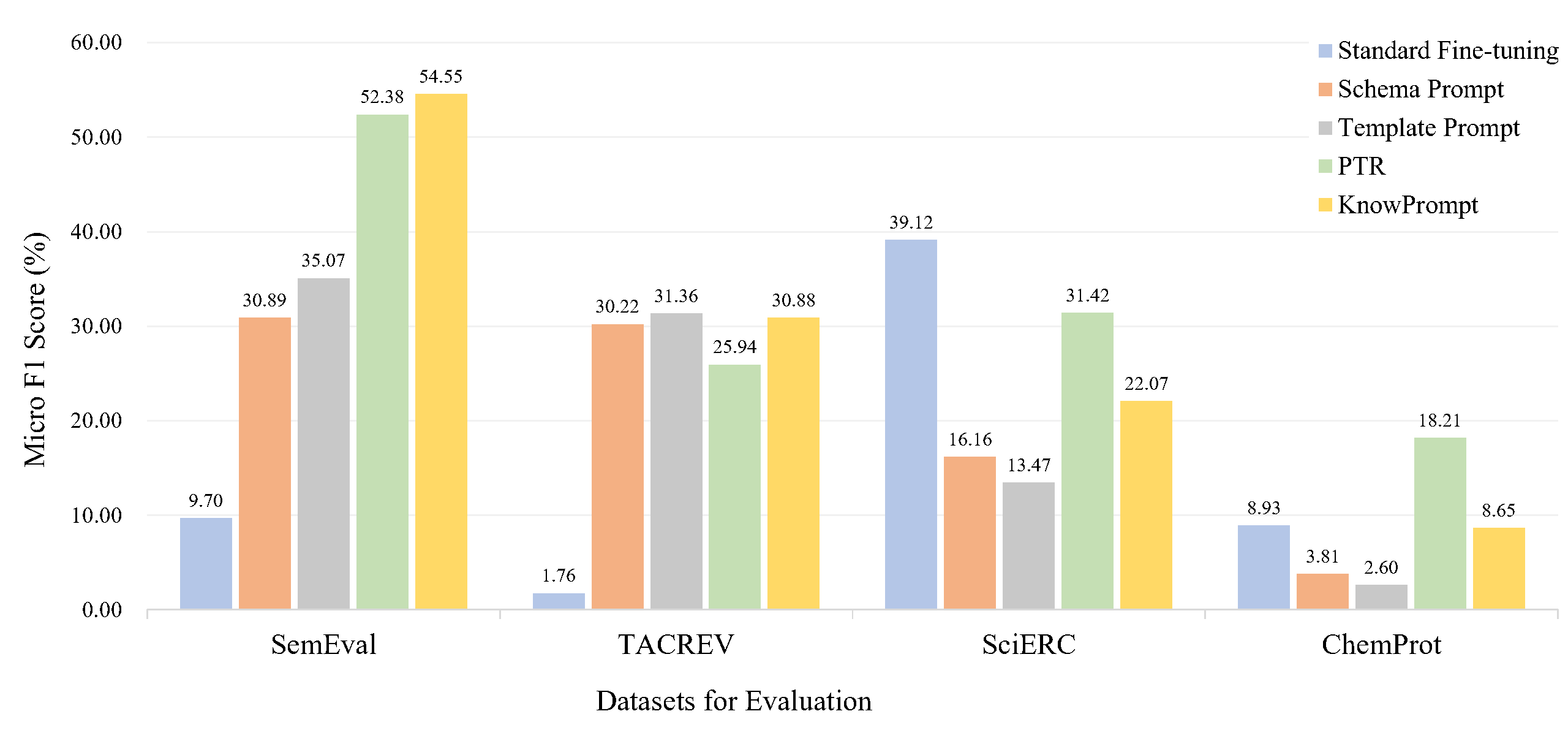}
    \caption{
    Micro F1 Scores (\%) of different prompts on 8-shot datasets. 
    \textit{RoBERTA-large} is used on SemEval and TACREV and \textit{BioBERT-large} is used on SciERC and ChemProt as backbone networks.}
    \label{fig:prompt}
\end{figure*}

\textbf{Finding 4:}
\textbf{RE systems struggle against difficulty in obtaining correct relations from cross-sentence contexts and among multiple triples.}
The extremely low F1 scores for 8-shot ChemProt, and DialogRE datasets in standard fine-tuning demonstrate this finding.
One text in ChemProt is related to too many relational triples (there are 347 texts related to 3 triples and 699 texts related to 2 triples in the training set).
At the same time, in DialogRE, the input text is extremely long (one text can contain 10 sentences).
Even with the powerful prompt-based tuning method, it is non-trivial to address the low-resource issue according to the unexpected \textcolor{red}{drop} in F1 scores of ChemProt and SciERC. 

\textbf{Finding 5: Self-training with unlabeled in-domain data may not always show an advantage for low-resource RE.}
There is much noise in those generated pseudo labels. 
Furthermore, for assigning labels in RE, both semantics and positions of entities in a text need to be considered, which is exceedingly challenging.
Therefore, the model with self-training cannot always obtain better performance in low-resource settings.



\subsection{Comprehensive Empirical Analysis}

\paragraph{Different Prompting Methods} 
To investigate the effects of different prompts, we conduct an empirical analysis on  SemEval, TACREV, SciERC and ChemProt as shown in Figure \ref{fig:prompt}.
We observe the following insights:
$(i)$ \textbf{Prompt-based tuning is more beneficial in general domains than specific domains for low-resource RE.}
Prompt-based tuning achieves the most gain, 44.85\% Micro F1 Score, by comparing fine-tuning and \textit{KnowPrompt} on 8-shot SemEval, while obtaining the worst drop, 25.65\% Micro F1 Score, by comparing fine-tuning and the template prompt on 8-shot SciERC even with the domain-specific PLM.
Except for the difficulty of these two datasets, general manual prompts have little domain knowledge related to vertical domains, hindering performance.
$(ii)$ \textbf{Entity type information in prompts is helpful for low-resource RE.} 
The head and tail entity types in prompts provide strong constraints between relations and their related entities.
Prompting methods with entity type information in \textit{KnowPrompt} and \textit{PTR} perform better than the template and schema-based prompt in most datasets, which illustrates that prompts with entity-type information are more appropriate for low-resource RE. 
The reason for the abnormal phenomenon that KnowPrompt and PTR obtain worse results than the template and schema-based prompts in TACREV is that annotation errors in the training set of TACREV \cite{DBLP:conf/aaai/StoicaPP21} can lead to overestimation of the performance of models depending on the side information of entities such as entity names, spans and types \cite{DBLP:journals/corr/abs-2102-01373}, and the templates of \textit{KnowPrompt} and \textit{PTR} are natural language sentences consisting of the head, and tail entities and their relations, which require high-quality annotated entity mentions, positions, types and relational words, while they are relatively trivial to the template and schema-based prompts.
\begin{table*}[t!]
  \centering
  \scalebox{0.67}{
    \begin{tabular}{lccccccccccccccccc}
    \toprule
    \multirow{3}[4]{*}{\textbf{Method}} & \multicolumn{8}{c}{\textbf{SemEval}}   &   & \multicolumn{8}{c}{\textbf{SciERC}} \\
\cmidrule{2-9}\cmidrule{11-18}      & \multicolumn{2}{c}{\textbf{Few}} & \multicolumn{2}{c}{\textbf{Medium}} & \multicolumn{2}{c}{\textbf{Many}} & \multicolumn{2}{c}{\textbf{Overall}} &   & \multicolumn{2}{c}{\textbf{Few}} & \multicolumn{2}{c}{\textbf{Medium}} & \multicolumn{2}{c}{\textbf{Many}} & \multicolumn{2}{c}{\textbf{Overall}} \\
      & MaF1 & MiF1 & MaF1 & MiF1 & MaF1 & MiF1 & MaF1 & MiF1 &   & MaF1 & MiF1 & MaF1 & MiF1 & MaF1 & MiF1 & MaF1 & MiF1 \\
    \midrule
    \textbf{Normal} & 50.42 & 74.58 & \textbf{89.53} & 89.02 & 90.17 & 90.59 & 83.40 & 90.01 &   & 69.98 & 67.78 & 88.05 & 87.52 & 92.98 & 91.93 & 83.27 & 89.01 \\
    \textbf{Re-sample} & 38.17 & 56.18 & 70.13 & 70.56 & 71.22 & 71.54 & 65.37 & 71.31 &   & 71.79 & 69.64 & 88.49 & 87.83 & 92.96 & 92.25 & 82.61 & 87.58 \\
    \textbf{DSC} & 49.80 & 73.87 & 87.84 & 88.00 & 88.97 & 89.52 & 82.19 & 89.00 &   & 71.57 & 69.90 & 89.94 & 89.51 & 93.51 & 92.88 & 83.09 & 88.91 \\
    \textbf{Focal} & 
    53.31 & 
    77.69 & 
    89.50 & 
    \textbf{89.57} & 
    \textbf{90.71} & 
    \textbf{91.06} & 
    \textbf{84.21} & 
    \textbf{90.55} & & 
    \textbf{73.47} & 
    \textbf{72.38} & 
    \textbf{91.88} & 
    \textbf{91.54} & 
    \textbf{94.83} & 
    \textbf{94.08} & 
    \textbf{84.83} & 
    \textbf{90.04} \\
    \textbf{GHM-C} & 00.00 & 00.00 & 3.39 & 6.27 & 70.42 & 75.81 & 43.79 & 70.99 &   & 71.34 & 69.28 & 89.42 & 88.82 & 93.90 & 93.33 & 82.95 & 88.81 \\
    \textbf{LDAM} & \textbf{53.53} & \textbf{79.66} & 88.71 & 88.98 & 90.32 & 90.60 & 83.83 & 90.15 &   & 72.32 & 70.55 & 88.48 & 87.73 & 94.61 & 93.98 & 83.31 & 89.22 \\
    \bottomrule
    \end{tabular}}

    \caption{F1 Scores (\%) on SemEval and SciERC datasets of diverse balancing methods via \textit{KnowPrompt}. \textit{MaF1} and \textit{MiF1} mean Macro F1 Score (\%) and Micro F1 Score (\%) respectively. \textit{Normal} means conducting the experiment without any balancing methods.}
      \label{tab:balance}
\end{table*}

\begin{table*}
  \centering
  \scalebox{0.65}{
    \begin{tabular}{lccccccrcccccc}
    \toprule
    \multicolumn{1}{c}{\multirow{3}[4]{*}{Method}} & \multicolumn{6}{c}{\textbf{SemEval}} &   & \multicolumn{6}{c}{\textbf{TACRED-Revisit}} \\
\cmidrule{2-7}\cmidrule{9-14}      & \multicolumn{3}{c}{30\%} & \multicolumn{3}{c}{100\%} &   & \multicolumn{3}{c}{30\%} & \multicolumn{3}{c}{100\%} \\
      & Context & Entity & All & Context & Entity & All &   & Context & Entity & All & Context & Entity & All \\
\cmidrule{1-7}\cmidrule{9-14}    WordNet's Synonym & \textbf{75.49} & 75.50  & \textbf{76.47} & \textbf{83.54} & \textbf{83.50} & \textbf{82.56} &   & 76.54  & \textbf{76.87} & 76.63  & \textbf{76.12} & \textbf{76.59} & \textbf{76.37} \\
    TF-IDF Similarity & 73.93  & \textbf{76.23} & 74.30  & 82.92  & 82.61  & 82.33  &   & \textbf{76.63} & 76.05  & \textbf{76.90} & 75.44  & 75.80  & 75.15  \\
    Contextual Word Embedding (RoBERTa) & 73.84  & - & 74.41  & 81.63  & - & 81.31  &   & 75.86  & 76.76  & 76.35  & 75.98  & 76.12  & 75.92  \\
    KnowPrompt (RoBERTa) & \multicolumn{6}{c}{69.90} &   & \multicolumn{6}{c}{77.00} \\
    \midrule
    \midrule
      & \multicolumn{6}{c}{\textbf{SciERC}} &   & \multicolumn{6}{c}{\textbf{ChemProt}} \\
\cmidrule{2-7}\cmidrule{9-14}    WordNet's Synonym & 77.70  & 76.98  & \textbf{77.54} & \textbf{79.36} & \textbf{79.40} & \textbf{79.92} &   & \textbf{57.37} & 57.56  & \textbf{57.03} & \textbf{53.36} & \textbf{57.11} & \textbf{54.27} \\
    TF-IDF Similarity & \textbf{78.50} & \textbf{77.33} & 73.92  & 78.30  & 79.38  & 79.38  &   & 41.22  & \textbf{58.26} & 47.95  & 43.06  & 54.60  & 43.63  \\
    Contextual Word Embedding (BioBERT) & 76.24  & 73.55  & 74.62  & 75.50  & 77.35  & 76.59  &   & 56.01  & 53.48  & 56.28  & 45.95  & 53.26  & 46.68  \\
    KnowPrompt (BioBERT) & \multicolumn{6}{c}{74.00} &   & \multicolumn{6}{c}{56.96 } \\
    \bottomrule
    \end{tabular}}
     \caption{
      Micro F1 Scores (\%) on four datasets generated by different data augmentation methods from 10\% training sets. Three DA methods are conducted to substitute words at three positions: only in contexts, only in entities and in both of them. ``-''  represents non-repeated data generated based on contextual word embedding is not available. 
      }
  \label{tab:DA}
\end{table*}

\paragraph{Different Balancing Methods}
\label{sec:balexp}
We also conduct experiments to validate the effectiveness of different balancing methods on two long-tailed datasets.
We categorize the classes into three splits based on the number of training instances per class, including \textit{Few}, \textit{Medium}, and \textit{Many}, and also report the results on the whole dataset with the \textit{Overall} setting in Table \ref{tab:balance} (split schemes are in Appendix \ref{app:split}).
We notice that with re-balancing methods (e.g., \textit{Focal Loss} and \textit{LDAM Loss}), the tail relations (Few) can yield better performance on both general and domain-specific datasets.
However, some technologies, such as \textit{GHM-C}, fail to contribute to performance gains. 
Overall, our empirical analysis illustrates that the RE performance can be improved with balancing methods, which indicates that long-tailed RE is a challenging classification task, and it should be paid more attention to developing suitable methodologies.

\paragraph{Different Data Augmentation} 
\label{sec:unlabel}
To evaluate the low-resource RE performance with more instances, we generate  30\% and 100\% augmented instances from 10\% training sets by substituting tokens based on three methods.
From Table \ref{tab:DA}, we notice that DA with WordNet can obtain the best performance improvement in most cases. 
Further, we observe that DA methods can rise by 13.6\% and 5.92\% Micro F1 Scores mostly on SemEval and SciERC compared to origin prompt-based tuning, demonstrating that DA contributes a lot in the low-resource scenario. 
Besides, we observe that the performance improvement is much smaller in specific domains, such as SciERC and ChemProt, than in the general domain.
We think that because there are many specific terms in vertical domains, it is challenging to obtain qualified augmented instances, which causes to yield lower performance improvement.

\section{Related Work}
\paragraph{General and Low-resource RE}
Relation extraction is essential in information extraction.
Learning algorithms for RE models involve feature-based methods \cite{kambhatla-2004-combining}, semi-supervised \cite{chen-etal-2006-relation, rosenfeld-feldman-2007-using, sun-etal-2011-semi}, graph-based methods \cite{zhang-etal-2018-graph, guo-etal-2019-attention, ijcai2020-505} and applies PLMs as the backbone \cite{lin-etal-2020-joint,DBLP:conf/ijcai/ZhangCXDTCHSC21,zheng-etal-2021-prgc,ttwuspeechre,chen-etal-2022-good,DBLP:conf/sigir/ChenZLDTXHSC22}.
Since labeled instances may be limited in practice, low-resource RE has appealed to researchers \cite{DBLP:journals/tacl/SaboEGD21}.

\paragraph{Prompting Methods for RE}
Though fine-tuning PLMs has waved the NLP community,  there is still a big gap between pre-training and fine-tuning objectives, hindering the few-shot performance.
Hence, prompt-based tuning is proposed in GPT-3 \cite{NEURIPS2020_1457c0d6} and drawn much attention. 
A series of researches have illustrate the decent performance of prompt-based tuning \cite{ autoprompt:emnlp20,  lester-etal-2021-power, li-liang-2021-prefix}, especially in few-shot classification tasks \cite{schick-schutze-2021-exploiting, liu2021gpt,DBLP:journals/corr/abs-2205-14704}. 
Typically, PTR \cite{han2021ptr} encodes prior knowledge using logic rules in prompt-based tuning with several sub-prompts for text classification. 
KnowPrompt \cite{DBLP:conf/www/ChenZXDYTHSC22} incorporates knowledge among relation labels into prompt tuning for RE with synergistic optimization for better performance.

\paragraph{Methods for Long-tailed Distribution Data}
Many re-balancing methods are proposed to tackle the long-tailed problem \cite{kang2019decoupling,nan-etal-2021-uncovering}.
Data distribution re-balancing methods re-sample the dataset into a more balanced data distribution \cite{DBLP:conf/icic/HanWM05, DBLP:conf/eccv/MahajanGRHPLBM18}.
Various re-weighing losses \cite{cui2019classbalancedloss, 10.1609/aaai.v33i01.33018577, DBLP:conf/acl/LiSMLWL20, DBLP:journals/pami/LinGGHD20,DBLP:conf/nips/CaoWGAM19} assign balanced weights to training samples from each class. 
For RE, \citet{nan-etal-2021-uncovering} introduces causal inference to mitigate the spurious correlation issues for information extraction.

\paragraph{Data Augmentation for NLP}
An effective method for NLP in low-resource domains is data augmentation. 
Token-level DA approaches include replacing tokens with their synonyms \cite{kolomiyets-etal-2011-model, wang-yang-2015-thats}, deleting  tokens \cite{iyyer-etal-2015-deep}, inserting random tokens \cite{wei-zou-2019-eda, DBLP:conf/www/MiaoLWT20} or replacing meaningless tokens with random tokens \cite{DBLP:conf/nips/XieDHL020, DBLP:conf/conll/NiuB18}.

\section{Conclusion}
We provide an empirical study on low-resource RE.
Specifically, we analyze the prompt-based tuning for few-shot RE, balancing methods for long-tailed RE datasets, and use data augmentation or unlabeled in-domain data.
We systematically evaluate baselines on 8 benchmark datasets in low-resource settings (e.g., 8-shot, 10\%) and provide insightful findings.
We hope this study can help inspire future research for low-resource RE with more robust models and promote transitioning the technology to real-world industrial scenarios.

\section{Limitations} 
With the fast development of low-resource RE, we cannot compare and evaluate all previous studies due to the settings and non-available open-sourced code.
Our motivation is to develop a universal, GLUE-like, and open platform on low-resource RE for the community.
We will continue to maintain the benchmark by adding new datasets.

\section*{Acknowledgment}

We would like to express gratitude to the anonymous reviewers for their kind comments. 
This work was supported by the National Natural Science Foundation of China (No.62206246, 91846204 and U19B2027), Zhejiang Provincial Natural Science Foundation of China (No. LGG22F030011), Ningbo Natural Science Foundation (2021J190), Yongjiang Talent Introduction Programme (2021A-156-G), Information Technology Center and State Key Lab of CAD\&CG, Zhejiang University.

\bibliography{anthology,custom}
\bibliographystyle{acl_natbib}

\appendix

\section{Implementation Details}
\label{app:exper}
\subsection{Settings}
\label{app:setting}
We detail the training procedures and hyperparameters for each of the datasets. 
We utilize PyTorch to conduct experiments with one NVIDIA RTX 3090 GPU.
All optimization is performed with the AdamW optimizer \cite{DBLP:conf/iclr/LoshchilovH19}.
The training is always continuous for 10 epochs without validation.
All pre-trained language models used in this work are downloaded from HuggingFace.
The names of PLMs are "hfl/chinese-roberta-wwm-ext-large'' for DuIE2.0 and CMeIE, "dmis-lab/biobert-large-cased-v1.1'' for SciERC and ChemProtm, and "roberta-large'' for other benchmark datasets.

\subsection{Prompting Methods}
In the prompt-based tuning experiments with \textit{KnowPrompt} (PyTorch-Lightning), the early stop in the original code is dropped.
The learning rate is set as $4e-5$ for all datasets.
Instead of using the original code for multi-labeled DialogRE with BCEloss, we implement experiments with DialogRE the same as the other seven datasets to unify our benchmark.

\subsection{Balancing Methods}
For re-sampling methods, we firstly use the sampler on all 10\% and 100\% imbalanced training sets to get nearly balanced training sets and then use them in all methods the same way as imbalanced datasets.
We leverage the official code of various re-weighting losses and provide the alternative parsing argument named "useloss'' for developers to choose them.

\subsection{Data Augmentation}
Different DA methods mentioned in \S \ref{sec:DA and ST} are utilized on English and Chinese datasets via \textit{nlpaug} and \textit{nlpcda}.
After generating augmented data, we merge them with original data in order to delete repeated instances that make no sense.
Then both original and augmented data are combined and fed into models to evaluate their performance.

\begin{table}[h]
  \centering
  \scalebox{0.9}{
    \begin{tabular}{lcc}
    \toprule
    \textbf{Relation} & \textbf{Number} & \textbf{Level} \\
    \midrule
    Other & 1145 & - \\
    \midrule
    Entity-Destination (e1,e2) & 686 & \multirow{11}[2]{*}{Many} \\
    Cause-Effect (e2,e1) & 536 &  \\
    Member-Collection (e2,e1) & 498 &  \\
    Entity-Origin (e1,e2) & 462 &  \\
    Message-Topic (e1,e2) & 399 &  \\
    Component-Whole (e2,e1) & 383 &  \\
    Component-Whole (e1,e2) & 382 &  \\
    Instrument-Agency (e2,e1) & 331 &  \\
    Product-Producer (e2,e1) & 321 &  \\
    Content-Container (e1,e2) & 304 &  \\
    Cause-Effect (e1,e2) & 280 &  \\
    \midrule
    Product-Producer (e1,e2) & 263 & \multirow{4}[2]{*}{Medium} \\
    Content-Container (e2,e1) & 135 &  \\
    Entity-Origin (e2,e1) & 121 &  \\
    Message-Topic (e2,e1) & 117 &  \\
    \midrule
    Instrument-Agency (e1,e2) & 79 & \multirow{3}[2]{*}{Few} \\
    Member-Collection (e1,e2) & 64 &  \\
    Entity-Destination (e2,e1) & 1 &  \\
    \bottomrule
    \end{tabular}}
    \caption{Relation splits on SemEval.}
  \label{tab:semevalsplit}
\end{table}

\begin{table}[h]
  \centering
    \scalebox{0.9}{
    \begin{tabular}{lcc}
    \toprule
    \textbf{Relation} & \textbf{Number} & \textbf{Level} \\
    \midrule
    USED-FOR & 1690 & \multirow{2}[2]{*}{Many} \\
    CONJUNCTION & 400 &  \\
    \midrule
    EVALUATE-FOR & 313 & \multirow{2}[2]{*}{Medium} \\
    HYPONYM-OF & 298 &  \\
    \midrule
    PART-OF & 179 & \multirow{3}[2]{*}{Few} \\
    FEATURE-OF & 173 &  \\
    COMPARE & 166 &  \\
    \bottomrule
    \end{tabular}}
  \caption{Relation splits on SciERC.}
  \label{tab:SciERCsplit}
\end{table}

\subsection{Self-training}
\label{app:self-train}
Given unlabeled data $\mathcal{D}^{\text{U}}$ and a few labeled data $\mathcal{D}^{\text{L}}$, we conduct self-training for semi-supervised learning.
The scheme is executed as the following steps \cite{huang-etal-2021-shot}:
\begin{enumerate}
    \item Train a teacher model $\Theta^{\text{T}}$ with gold-labeled data $\mathcal{D}^{\text{L}}$ via cross-entropy.
    \item Use the trained teacher model $\Theta^{\text{T}}$ to generate soft labels on unlabeled data $\mathcal{D}^{\text{U}}$:
    \begin{equation}
        \tilde{y}_i=f_{\Theta^{\text{T}}} (\tilde{x}_i), \quad \tilde{x}_i \in \mathcal{D}^{\text{U}}
    \end{equation}
    \item Train a student model $\Theta^{\text{S}}$ via cross-entropy $\mathcal{L}$ on both gold-labeled data $\mathcal{D}^{\text{L}}$ and soft-labeled data $\mathcal{D}^{\text{SU}}$. The loss function of $\Theta^{\text{S}}$ is:
    \begin{align}
        \begin{split}
            \mathcal{L}_{\texttt{STU}} =
              & \frac{1}{\left| \mathcal{D}^{\text{L}} \right|} \sum_{x_i \in \mathcal{D}^{L}} \mathcal{L}(f_{\Theta^{\text{S}}}(x_i),y_i) \\
              & + \frac{\lambda_{\text{U}}}{\left| \mathcal{D}^{\text{U}} \right|} \sum_{\tilde{x}_i \in \mathcal{D}^{U}} \mathcal{L}(f_{\Theta^{\text{S}}}(\tilde{x}_i),\tilde{y}_i)
        \end{split}
    \end{align}
\end{enumerate}
where $\lambda_{\text{U}}$ is the weighting hyper-parameter, and we set it $0.2$ in this work. 
It is an alternative to iterate from Step 1 to Step 3 multiple times by initializing $\Theta^{\text{T}}$ in Step 1 with newly learned $\Theta^{\text{S}}$ in Step 3. We only perform self-training once in our experiments for simplicity because the result is not good, and it is not sensitive to continue the next iteration.

\section{Class Splits in Balancing Methods Evaluation}
\label{app:split}
The few-level, medium-level and many-level relation splits based on the number of each relation class on SemEVal and SciERC are shown in Table \ref{tab:semevalsplit} and Tabel \ref{tab:SciERCsplit} for comparative experiments on different re-weighting losses in \S \ref{sec:balexp}.
\end{document}